\documentclass[10pt,twocolumn,letterpaper]{article}

\usepackage[pagenumbers]{cvpr} 

\usepackage{graphicx}
\usepackage{amsmath}
\usepackage{amssymb}
\usepackage{booktabs}

\usepackage{latexsym}
\usepackage{bm}

\usepackage[pagebackref,breaklinks,colorlinks]{hyperref}

\usepackage[capitalize]{cleveref}
\crefname{section}{Sec.}{Secs.}
\Crefname{section}{Section}{Sections}
\Crefname{table}{Table}{Tables}
\crefname{table}{Tab.}{Tabs.}

\newcommand{\myparagraph}[1]{\textbf{#1 ---}}
\newcommand\mydots{\makebox[0.7em][c]{.\hfil.\hfil.}}

\def\cvprPaperID{*****} 
\def\confName{CVPR}
\def\confYear{2022}

\begin{document}

\title{An experimental study of the vision-bottleneck in VQA}

\author{Pierre Marza\\
Orange Innovation, Cesson-Sévigné, France\\
LIRIS, INSA Lyon, UMR CNRS 5205, France\\
{\tt\small pierre.marza@insa-lyon.fr}
\and
Corentin Kervadec\\
Orange Innovation, Cesson-Sévigné, France\\
LIRIS, INSA Lyon, UMR CNRS 5205, France\\
{\tt\small corentin.kervadec@orange.com}
\and
Grigory Antipov\\
Orange Innovation, Cesson-Sévigné, France\\
{\tt\small grigory.antipov@orange.com}
\and
Moez Baccouche\\
Orange Innovation, Cesson-Sévigné, France\\
{\tt\small moez.baccouche@gmail.com}
\and
Christian Wolf\\
LIRIS, INSA Lyon, UMR CNRS 5205, France\\
{\tt\small christian.wolf@insa-lyon.fr}
}
\maketitle

\begin{abstract}
As in many tasks combining vision and language, both modalities play a crucial role in Visual Question Answering (VQA). To properly solve the task, a given model should both understand the content of the proposed image and the nature of the question. While the fusion between modalities, which is another obviously important part of the problem, has been highly studied, the vision part has received less attention in recent work. Current state-of-the-art methods for VQA mainly rely on off-the-shelf object detectors delivering a set of object bounding boxes and embeddings, which are then combined with question word embeddings through a reasoning module. In this paper, we propose an in-depth study of the vision-bottleneck in VQA, experimenting with both the quantity and quality of visual objects extracted from images. We also study the impact of two methods to incorporate the information about objects necessary for answering a question, in the reasoning module directly, and earlier in the object selection stage. This work highlights the importance of vision in the context of VQA, and the interest of tailoring vision methods used in VQA to the task at hand.
\end{abstract}

\section{Introduction}
\label{sec:introduction}
\noindent
Visual Question Answering (VQA) \cite{antol2015vqa}, which requires answering a textual question on an image, has become an interesting test-bed for the evaluation of the reasoning and generalization capabilities of trained computational models. It deals with open questions and large varieties, and solving an instance can involve visual recognition, logic, arithmetic, spatial reasoning, intuitive physics, causality and multi-hop reasoning. It also requires combining two modalities of different nature: images and language. The visual modality in particular is of high dimension and needs to be compressed and reduced, focusing attention on the information most relevant to the underlying reasoning problem. This also necessitates identifying key regions or objects, and relating them to each other and to the question. The latter is done in later stages in current models.

Whereas early work on VQA focused on the multi-modal fusion aspects and represented images as grid-shaped feature maps \cite{yang2016stacked,kim2018bilinear,ben2017mutan,ben2019block}, current State-Of-The-Art (SOTA) models \cite{Tan2019LXMERTLC, zhang2021vinvl, li2020oscar} reason over a set of objects obtained via a pretrained detector as introduced in UpDn \cite{Anderson_2018_CVPR}, or more recently VinVL \cite{zhang2021vinvl}. Feature embeddings and Bounding Box (BB) coordinates are fused with word embeddings extracted from the question, for instance with intra- and cross-modality attention \cite{Yu_2019_CVPR,Tan2019LXMERTLC,kervadec2019weak}. The two different input modalities, image and language, are thus dealt with independently in the initial stages, and combined only after each one has been transformed into a suitable high-level representation. 

Recently it has been shown, that Oracle models with perfect sight, working on ground truth visual input, not only perform better in absolute terms, but also that they are less prone to exploit spurious biases and short-cuts in the training data \cite{ReasoningCVPR2021,ReasoningNeurIPS2021}. This provides evidence that the difficulty in learning vision and language reasoning is in large part caused by the computer vision bottleneck and its deficiencies, leading to uncertainties and noise.

In this work, we study the impact of diverse characteristics of such visual objects extracted by a baseline detector on the downstream VQA performance. More specifically, we consider the consequence of varying both the quantity and the quality, i.e., precision of bounding box (BB) coordinates and semantic description, of visual objects fed to VQA models. We then explore the impact of additional task-related information during training, both on the object detection side to see if the quality of selected visual objects can be improved and/or if it is possible to reduce the quantity of objects without degrading performance, and on the side of the VQA model to study if guiding it to differentiate between relevant and useless visual information regarding the asked question impacts downstream performance.

\myparagraph{Quantity of visual objects} Currently, the top-$k$ detected BBs are kept based on a classification score reflecting the confidence of the model in the semantic class of the object within the BB. We study the impact of varying the number $k$ of BBs fed to a VQA model. If $k$ is small, objects necessary for reasoning about the question might be missing, simply because $k$ is smaller than the number of ground truth objects necessary for reasoning and/or because ranking of BBs is done without taking the asked question into account. Visual objects with the highest score are indeed not the most useful to answer the question, but the most easily recognizable from an object detection viewpoint. This also means that a high $k$ could lead to lots of unnecessary objects that might overwhelm the VQA system. 

\myparagraph{Quality of visual objects} We further analyze the importance of feeding the exact $(x,y)$ coordinates of the BBs within the image, as well as the correct embedding describing the content of the image at this location to a VQA model. More generally, we also emulate a perfect detection of objects necessary to answer the asked question, showcasing a strong impact on the downstream performance. A lot has thus still to be done in Vision for VQA and such contributions might have a non-negligible impact.

\myparagraph{Language-grounded object selection} We study a way we could directly act at the level of the detector, by adding a language-grounded object selection module to the Faster R-CNN detection framework \cite{faster_rcnn2015}, aiming to select objects necessary for reasoning (Figure~\ref{fig:teaser}b). We thus consider the impact of replacing the classical scoring of proposed BBs based on the confidence of the classifier (see Section ~\ref{subsec:object_detection} for more details about object detection), with a task-related scoring approach that is dependent on the question. This could have two advantages, that we study in this work. Firstly, it can filter out candidate objects which are not necessary for reasoning. Secondly, and more importantly, it could retrieve additional objects, which would have been missed by a pure bottom-up process, in particular due to domain shifts in pre-training and without any contextual information from the asked question.
Another question is whether language grounding can not only retrieve objects directly referenced in the question, but also objects required for correct reasoning but not directly addressed, for instance the BB of a woman corresponding to the answer of the question ``\emph{Who is wearing the sweater?}''. We experimentally show that modern Transformer \cite{NIPS2017_7181}-based language models are indeed capable of providing the necessary context for the retrieval of indirectly referenced objects.

\myparagraph{Additional supervision of the VQA reasoning model} We finally explore a method to help VQA models differentiating between useful and unrelated objects extracted by a detector. This takes the form of an auxiliary supervision, and more precisely a binary classification problem where the VQA model, while also supervised to answer properly the asked question, must predict if visual objects given as input are necessary to reason. This is interesting, knowing that pure bottom-up detectors tend to extract most of the objects within images, while only a fraction is indeed necessary to answer the question.

Both language grounding and auxiliary supervision of the VQA model benefit from the Ground Truth (GT) annotations on the objects necessary for reasoning of a given image-question pair available for the GQA dataset \cite{Hudson2019GQAAN}. 

\section{Related work}
\label{sec:relatedwork}
\myparagraph{Visual Question Answering (VQA)}
as a task was introduced together with various datasets, such as VQAv1 \cite{antol2015vqa} and VQAv2 \cite{goyal2017making} (built from human annotators), or CLEVR \cite{johnson2017clevr} and GQA \cite{Hudson2019GQAAN}.
Additional splits were proposed to evaluate specific reasoning capabilities.
For instance, VQA-CP \cite{vqa-cp} explicitly inverts the answer distribution between train and test splits, whereas GQA-OOD \cite{RosesCVPR2021} focuses on rare (Out-Of-Distribution) question-answer pairs, and shows that many VQA models strongly rely on dataset biases.
While an exhaustive survey of VQA approaches is out of the scope of this paper, one can mention some major works, including approaches based on attention networks \cite{yang2016stacked}, object attention mechanisms \cite{Anderson_2018_CVPR}, language-vision co-attention \cite{lu2016hierarchical}, bilinear fusion \cite{kim2018bilinear}, tensor decompositions \cite{ben2017mutan,ben2019block}, neural-symbolic reasoning \cite{yi2018neural}, probabilistic graphs \cite{hudson2019learning} and, more recently, Transformer-based models \cite{Yu_2019_CVPR,Tan2019LXMERTLC,kervadec2019weak, zhang2021vinvl}.

\myparagraph{Object detection}
Object detection has been an active research area in computer vision for decades, due to its wide range of applications. It is a challenging task, since it combines two problems: distinguishing foreground objects from background through localizing their BBs and predicting their class labels. With the development of deep learning, object detection performance has been continuously improving, both in efficiency and accuracy. Deep learning-based approaches for object detection can be roughly divided into two categories, depending on whether the detection is performed in one or two stages. The first category, which includes well-known detectors like SSD \cite{liu2016ssd}, RetinaNet \cite{lin2017focal}, different versions of YOLO \cite{redmon2016you, redmon2017yolo9000, redmon2018yolov3}, and the recent DETR Transformers-based detector \cite{carion2020end}, directly predicts object BBs from the input image. On the other hand, two-stage approaches combine a region proposal algorithm or model with a region-wise refinement stage. Within the latter category, different improvements over the well-known R-CNN detection framework \cite{ girshick2014rich} have been proposed \cite{girshick2015fast,faster_rcnn2015,cai2019cascade}. In particular, Faster R-CNN \cite{faster_rcnn2015} is widely adopted as having a good trade-off between accuracy and efficiency. In this work, as in many recent VQA approaches \cite{Anderson_2018_CVPR,Yu_2019_CVPR,Tan2019LXMERTLC}, we use a ResNet CNN within a Faster R-CNN detection framework \cite{faster_rcnn2015}.

\myparagraph{Object detection in VQA}
While being very different, many recent VQA approaches share the same global scheme, in which the two modalities, i.e., image and question, are first processed separately and then fused to predict the answer. On the visual side, early work \cite{yang2016stacked,ben2017mutan} used grid-like feature maps extracted from a CNN as input features. However, as pointed out by \cite{teney2018tips}, these grid features have been gradually replaced by bottom-up attention approaches \cite{Anderson_2018_CVPR}, which provide object-specific features. Thus, almost all modern VQA approaches \cite{kim2018bilinear, Yu_2019_CVPR, Tan2019LXMERTLC} use bottom-up attention with off-the-shelf object detectors, especially the well-established Faster R-CNN \cite{faster_rcnn2015}.
Recently, \cite{zhang2021vinvl} introduces VinVL, an improved object-level visual representation for vision and language tasks. It is also based on the Faster RCNN~\cite{faster_rcnn2015} object detector, but trained with a ResNeXt-152 C4 backbone, and on a bigger dataset.
Nevertheless, \cite{jiang2020defense} is questioning this dominant choice, by showing that feeding a VQA system with all detected objects’ embeddings, without selecting which ones are relevant to answer the question, is equivalent to using grid features calculated on the entire image. This suggests that current VQA systems do not fully exploit the high object localization ability of the Faster R-CNN detector.
Our work is related, as it further studies the impact of diverse characteristics of extracted visual objects, as well as shows how higher quality question-related visual inputs in the form of ground-truth objects improve significantly the VQA performance.
Finally, our work has also connections with the recent MEDTR~\cite{kamath2021mdetr}, which proposes to jointly train the object detection module with the language modality using alignment losses. However, it differs from our study as the object detection is based on a Transformer architecture (following DETR~\cite{carion2020end}). In addition, our experimental study stands for its analysis of the impact of the quantity and quality of objects detected on the visual reasoning ability, measured using VQA accuracy.

\myparagraph{Links with language-grounded vision}
Without being directly related to language-grounded vision (also known as \emph{visual grounding}, \cite{ karpathy2015deep}), our work falls into a category of recent work in VQA, which study the use of language grounding as an additional supervision signal. Unlike other VQA approaches, which consider attention maps as internal latent variables, and which are not learned with supervision, these methods include a supervised language-grounded attention scheme in the VQA model. The benefit of using such supervision signal has been raised by \cite{das-etal-2016-human}, who collected the VQA Human Attention Dataset (VQA-HAT), and concluded that humans use different visual cues than VQA models to answer a given question. Following this claim, \cite{gan2017vqs} proposed an approach to identify the image regions used by humans to answer visual questions, and used them to train a deep learning model to predict these language-grounded attention maps. The authors showed that combining them with a standard VQA model slightly improves performance. \cite{selvaraju2019taking} proposed an approach for constraining the sensitivity of deep networks to specific input visual regions, which are considered by humans as important to answer a given question. Other work by \cite{qiao2018exploring} and \cite{zhang2019interpretable} uses the VQA-HAT dataset \cite{das-etal-2016-human} to train a language-grounded attention proposal network, and integrate its predictions to supervise attention in VQA systems. All these efforts led to improvements in VQA performance.

To some extent, our approach has connections with all these works since we also propose to use language-grounded attention supervision, either in the VQA model directly, or integrated into the detection stage, thus separated from the global supervision signal (i.e., the answer to the question).
A difference is that we used object/word alignments of the recent GQA dataset \cite{Hudson2019GQAAN} instead of the human attention annotation of the VQA-HAT dataset \cite{das-etal-2016-human} (see section~\ref{subsec:gqa_dataset}). This choice is mainly due to the small number of annotated training samples of the latter: about $13\%$ of the VQA-v2 dataset \cite{goyal2017making}.
 
\section{Preliminaries and experimental setup}
\noindent
We first introduce the necessary methodology on Visual Question Answering (VQA) and the corresponding computer vision pipelines required for understanding the experimental contributions in the following sections.

\subsection{Object detection in VQA}
\label{subsec:object_detection}
\noindent
In the context of VQA, a standard object detection model $\phi_d(\cdot)$ can be summarized as a function taking as input an image $\bm{I}$ and returning a set of feature embeddings $\bm{f}{=}\{\bm{f}_0,\bm{f}_1,\mydots,\bm{f}_{N_O-1}\}$ and the respective bounding box coordinates $\bm{b}=\{\bm{b}_0,\bm{b}_1,\mydots,\bm{b}_{N_O-1}\}$ constituting the resulting set $\bm{o}$ of $N_O$  objects:
\begin{equation}
    \bm{o}=(\bm{f},\bm{b})=\phi_d(\bm{I})
\end{equation}
which are fed into a VQA model. Most recent VQA systems rely on two-stage object detectors, whose first stage $\phi_r(\cdot)$ proposes potential candidate regions $\bm{r}{=}\{(\bm{f}_0,\bm{b}_0),(\bm{f}_1,\bm{b}_1),\mydots,(\bm{f}_{N_R-1},\bm{b}_{N_R-1})\}$, which are then further confirmed or rejected by a second-stage classification module $\phi_c(\cdot)$:
\begin{equation}
    \bm{o}=\phi_d(\bm{I})=\phi_c(\phi_r(\bm{I}))
\end{equation}
The predictor $\phi_r(\cdot)$ can be a dedicated neural network \cite{faster_rcnn2015} called Region Proposal Network (RPN). The main objective of the classification module $\phi_c(\cdot)$ is to classify each region proposal as one of $N_{C}$ predefined classes. Depending on the implementation, it may also refine the $N_R$ BBs, or choose to keep the original positions proposed by $\phi_r(\cdot)$.
It also delivers a confidence score $\phi_{cs}(\bm{l})$ for each proposal $\bm{l}\in \bm{r}$, which usually serves as a criterion to select them based on a threshold $\theta_{c}$:
\begin{equation}
    \forall \bm{l}\in\bm{r}: \phi_c(\bm{l}) \in \bm{o} \textrm{ if }  \phi_{cs}(\bm{l})>\theta_{c}
\end{equation}
Following a large body of work in VQA (\textit{cf.} Section~\ref{sec:relatedwork}), we choose Faster R-CNN \cite{faster_rcnn2015} as a baseline $\phi_d$ for our experiments.
In our experiments we reproduce the implementation of \cite{Anderson_2018_CVPR}, which is popular in VQA due to its multitask setting with, both, object and attribute classification objectives. However, our contribution can be easily extended to other implementations, such as the one recently proposed in VinVL~\cite{zhang2021vinvl}.

\subsection{Bottom-Up-Top-Down (UpDn)}
\label{subsec:updn}
\noindent
To evaluate the impact on downstream VQA performance of diverse factors, as well as supervision signals, all our experiments are done with the Bottom-Up-Top-Down (UpDn) \cite{Anderson_2018_CVPR} model, which is a simple and yet efficient model for VQA, making it an interesting baseline in the context of this study.

\myparagraph{UpDn Model} The question is first encoded as the hidden representation of a GRU \cite{cholearning}, fed with word embeddings sequentially. $k$ visual inputs are extracted by an object detector, as presented in the previous subsection. This process is purely bottom-up as it is not conditioned on the asked question. Then, a standard top-down mechanism is applied to compute a weight for each visual region, related to its relevance to the question. This is achieved through a concatenation of both modalities, followed by linear layers, to finally produce a distribution over the $k$ BBs with a softmax operator. The latent representation of the question is finally fused with the weighted sum of all visual inputs through element-wise multiplication, to serve as input to linear layers outputting a distribution over possible answers. The answer with the highest probability is picked.

\myparagraph{Training UpDn}
UpDn is trained during 5 epochs with a batch size of $512$ using the Adam optimizer. At the beginning of the training, we linearly increase the learning rate from $2e^{-3}$ to $2e^{-1}$ during 3 epochs, followed by a decay by a factor of $0.2$ at epochs 10 and 12. We use publicly available implementations at \footnote{\url{https://github.com/MILVLG/openvqa}}.

\subsection{Transformers}
\label{subsec:transformers}
We provide a brief overview of Intra-modality and Cross-modality Transformers, which will be involved in one of our studies about grounding object detection with language.

\myparagraph{Intra-modality Transformers}
are self-attention modules similar to~\cite{NIPS2017_7181}. Given an input sequence $\bm{x}~=~(\bm{x_1},\dots,\bm{x_n})$ of the embeddings of the same length $d$, they calculate an output sequence:
\begin{equation}
    \bm{x'}=t_{-}(\bm{x})=\sum_{j}\bm{\alpha}_{ij}\bm{x}^v_{j},
\end{equation}
by defining the query $\bm{x}^q$, key $\bm{x}^k$ and value $\bm{x}^v$ vectors which are calculated with the respective trainable matrices $\bm{x}^q=\bm{W}^{q}\bm{x}$, $\bm{x}^k=\bm{W}^{k}\bm{x}$ and $\bm{x}^v=\bm{W}^{v}\bm{x}$. In particular, $\bm{x}^q$ and $\bm{x}^k$ are used to calculate the self-attention weights $\bm{\alpha}_{\cdot j}$ as following: $\bm{\alpha}_{\cdot j}=(\bm{\alpha}_{1j},\dots,\bm{\alpha}_{ij},\dots,\bm{\alpha}_{nj})=\sigma(\frac{{\bm{x}^{q}_{1}}^{T}\bm{x}^{k}_{j}}{\sqrt{d}},\dots,\frac{{\bm{x}^{q}_{i}}^{T}\bm{x}^{k}_{j}}{\sqrt{d}},\dots,\frac{{\bm{x}^{q}_{n}}^{T}\bm{x}^{k}_{j}}{\sqrt{d}})$, with $\sigma$ being the softmax operator.

\myparagraph{Cross-modality Transformers} are defined as in~\cite{Tan2019LXMERTLC} and,
given a pair of embeddings of different modalities $(\bm{x} ,\bm{y})$, they consist of a quadruplet of intra-modality Transformers $t_{-}^{\bm{x}\rightarrow \bm{y}}$, $t_{-}^{\bm{x}\leftarrow \bm{y}}$, $t_{-}^{\bm{x}}$ and $t_{-}^{\bm{y}}$, which perform cross-attention from one modality to the other through different assignments of key, query and value functions over $\bm{x}$ and $\bm{y}$. 
More precisely, the input embeddings $\bm{x}$ and $\bm{y}$ are firstly processed by the transformers $t_{-}^{\bm{x}\leftarrow \bm{y}}(\cdot,\cdot)$ and $t_{-}^{\bm{x}\rightarrow \bm{y}}(\cdot,\cdot)$, respectively, which guide the attention on one modality via the embeddings of the opposite modality.
After that, the resulting sets $\bm{x}'$ and $\bm{y}'$ are passed through the simple intra-modality transformers $t_{-}^{\bm{x}}(\cdot)$ and $t_{-}^{\bm{y}}(\cdot)$ to obtain the output sets of embeddings $\bm{x}''$ and $\bm{y}''$.
The details are presented in equations below:
\begin{equation}
  \left\{
      \begin{aligned}
        \{(\bm{x_i}'',\bm{y_i}'')\}&=t_{\times}(\bm{x},\bm{y})=(t_{-}^{\bm{x}}(\bm{x}'),t_{-}^{\bm{y}}(\bm{y}')) \\
        \bm{x}'&=t_{-}^{\bm{x}\leftarrow \bm{y}}(\bm{x},\bm{y})=t_{-}^{\bm{x}\leftarrow \bm{y}}(\bm{x}^q,\bm{y}^k,\bm{y}^v) \\
        \bm{y}'&=t_{-}^{\bm{x}\rightarrow \bm{y}}(\bm{x},\bm{y})=t_{-}^{\bm{x}\rightarrow \bm{y}}(\bm{y}^q,\bm{x}^k,\bm{x}^v) \\
      \end{aligned}
    \right.
\end{equation}

\subsection{GQA dataset}
\label{subsec:gqa_dataset}
\noindent
We conduct all experiments on the challenging GQA dataset\footnote{\url{https://cs.stanford.edu/people/dorarad/gqa/about.html}} \cite{Hudson2019GQAAN}, as it is well suited for experimenting with the impact of visual objects due to its rich annotations, while remaining a challenging benchmark for recent VQA methods.
The questions in GQA have been generated algorithmically based on humanly annotated scene graphs for the respective images.
This property guarantees the availability of the Ground Truth (GT) annotations for the image objects required to answer each VQA question.

The following VQA metrics are used in the experiments: (i) classification \textit{Accuracy}, (ii) \textit{\# of selected objects per question} fed to VQA models during training and testing (useful for evaluation of our object selection module), and two secondary metrics which were proposed with the GQA dataset \cite{Hudson2019GQAAN}, namely: (iii) \textit{Binary} questions and (iv) \textit{Open} questions.

\begin{figure}[t] \centering
     \includegraphics[width=1\linewidth]{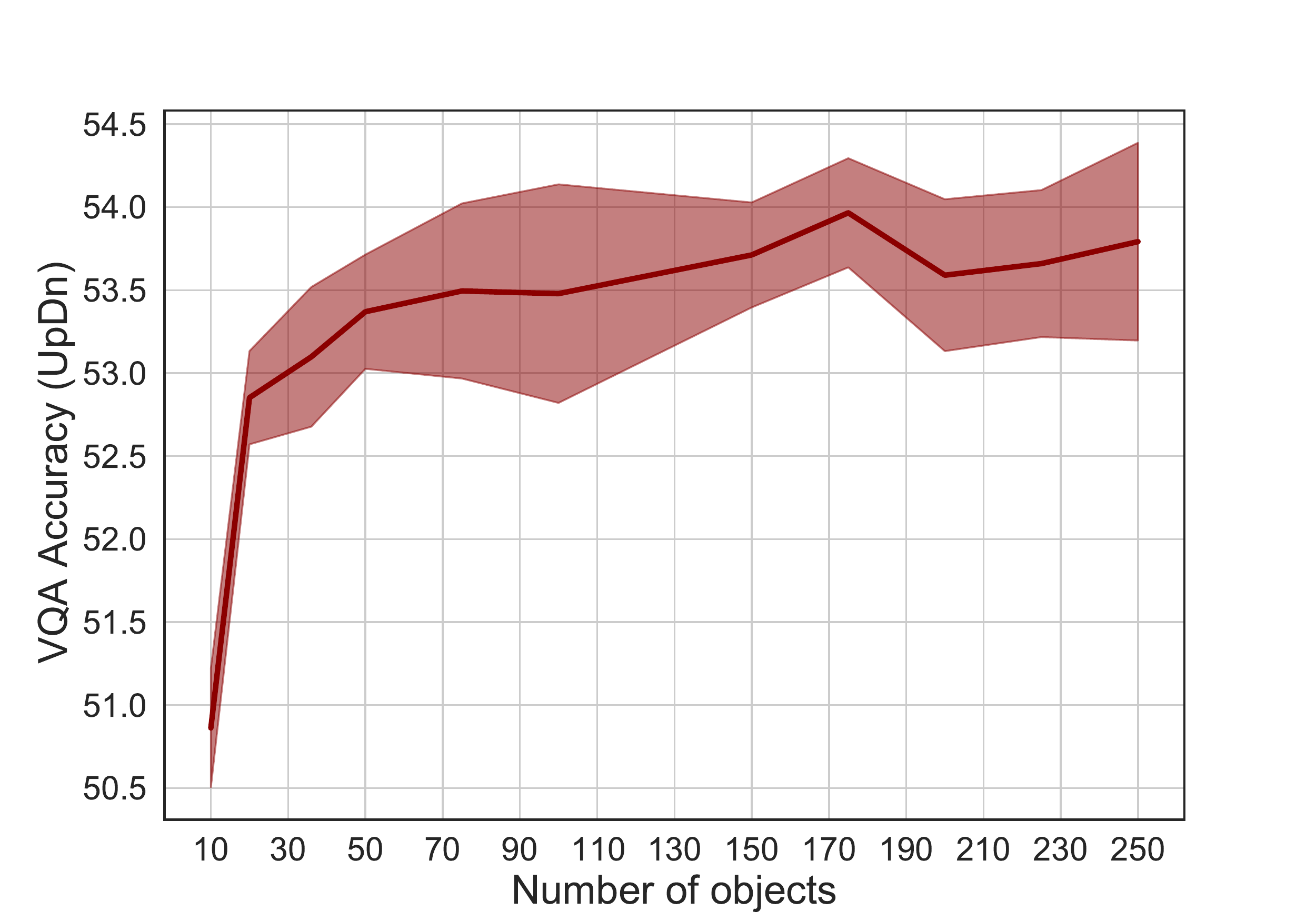}
     \caption{Impact of the number of objects retrieved by Faster R-CNN on downstream VQA performance obtained by the UpDn model on the test-dev set (solid curve, mean of 10 training runs per point, standard deviation shaded).
    Passed 50 objects, the system becomes unstable (high variance), and beyond 175 objects, accuracy tends to stagnate.
     \label{fig:impact_nb_objects_vqa}}
 \end{figure}

\section{The computer vision bottleneck in VQA}
\label{sec:cvbottleneck}
\noindent
Drawing conclusions from recent work providing evidence for the computer vision bottleneck in vision and language reasoning \cite{ReasoningCVPR2021, ReasoningNeurIPS2021}, we delve into details and provide an experimental study on the impact of quality of object detection on VQA. We look at the problem from multiple angles, and we study the impact of the number of objects processed by the reasoning module  and the impact of detection quality.

\subsection{The impact of object quantity}
\noindent
We explore the question whether visual inputs provided by baseline object detectors can limit the performance of VQA systems due to either too few, or an overwhelming quantity of proposed visual objects. While modern VQA models can themselves filter unnecessary detections, it is not clear whether this has an impact on reasoning performance.

To this end, we train and evaluate the VQA model UpDn \cite{Anderson_2018_CVPR} on the GQA train+val and test-dev sets respectively, while imposing a hard threshold on the number of objects detected with the Faster R-CNN baseline during both phases (the same maximum number of visual objects is used during training and validation). We thus report a plot of VQA accuracy as a function of the number of input visual objects (average of 10 different training runs), shown in Fig. \ref{fig:impact_nb_objects_vqa}. As expected, the VQA model is logically impacted by very low numbers of objects, since  objects required for reasoning might not be selected. 

Performance also slightly drops if too many objects are fed into the VQA module. Beyond 50 objects, VQA performance becomes quite unstable with large variations, and passed 175 objects, performance tends to stagnate. There is evidence that downstream VQA models are disturbed if they are flooded with too much information.

\begin{table*}[t] \centering
{\small
\begin{tabular}{ c c c c c c c }
\multicolumn{7}{c}{} \\
\toprule
\textbf{GT BB} & \textbf{GT embeddings} & \textbf{Perturbed} & \textbf{Perturbed} & \textbf{Accuracy} & \textbf{Binary} & \textbf{Open} \\
\textbf{boxes} & \textbf{(1-in-K class)} & \textbf{B. boxes} & \textbf{Embeddings} &
 &  &  \\ 
\midrule
${-}$ & ${-}$ & ${-}$ & ${-}$ &${60.24}$ & ${73.57}$ & ${47.75}$ \\
${\checkmark}$  & ${-}$ & ${\checkmark}$ & ${\checkmark}$ & ${59.58}$ & ${76.75}$ & ${43.48}$ \\
${\checkmark}$  & ${-}$ & ${\checkmark}$ & ${-}$ & ${69.21}$ & ${82.15}$ & ${57.08}$ \\
${\checkmark}$  & ${-}$ & ${-}$ & ${-}$ & ${69.21}$ & ${82.18}$ & ${57.06}$\\
${\checkmark}$  & ${\checkmark}$ & ${-}$ & ${-}$ & ${83.29}$ & ${82.93}$ & ${83.62}$\\
\bottomrule
\end{tabular}
}
\caption{\label{tab:results_oracles_perturbation}Impact of object detection quality (embeddings and BBs) on the UpDn VQA model, evaluated as comparison with oracles on GQA balanced validation set.} 
\end{table*}

\subsection{The impact of object detection quality}
\noindent
We evaluate the general impact of the quality of an object detection system on VQA performance.
In particular, we estimate the upper bound of the VQA model performance, by evaluating it with an oracle object detector, namely: (i) with GT bounding boxes (visual objects necessary to answer the question) instead of the regressed ones, and (ii) with GT embeddings in the form of the GT class one-hot vectors, instead of the detector's embeddings. These oracles simulate a perfect detector that would only output the set of visual objects necessary to reason about the asked question.
All tests are performed with standard Faster R-CNN and the UpDn VQA model trained  and evaluated on the balanced train and validation sets of the GQA dataset \cite{Hudson2019GQAAN} respectively.

Table~\ref{tab:results_oracles_perturbation} shows that more than $20$ VQA accuracy points are gained when both features and bounding boxes are taken from the oracle detector.
This confirms our intuition that there is a large room for improvement on the object detection side of VQA.
Moreover, analysing the gain brought by perfect selection of bounding boxes alone, one may notice that it can bring more than $+9$ pts of improvement for VQA.
This result seems to be particularly important, as it showcases how properly selecting necessary objects is essential. 
It corroborates the interest of improving the visual recognition pipeline for VQA, in particular improvements on the recall of objects required for reasoning, and the quality of object embeddings. 

We also measure to what extent the regression of the exact BB coordinates is essential for VQA, evaluating the scores under the perturbation of the GT coordinates.
For each GT bounding box coordinate, we sample random translations from a uniform distribution over $[-\frac{l}{2};+\frac{l}{2}]$, where $l$ is the size of the BB along the axis at hand. The results are shown in the 3rd row of Table~\ref{tab:results_oracles_perturbation} and paint a clear picture: given the rather strong amplitude of the coordinate perturbations, the drop in performance is surprisingly small.
On the contrary, if in addition to the bounding box coordinate perturbations, we also input the feature embeddings extracted by the detector for the perturbed location within the image to the VQA model, the performance drastically drops (2nd row in Table~\ref{tab:results_oracles_perturbation}).
This corroborates the intuition that answering questions in current applications and datasets requires a rather coarse knowledge of where objects are mostly restricted to their spatial relationships with other objects (\textit{left, right, above, under, below}, etc.), but a quite precise knowledge of the type of objects involved is necessary.
In other words, it is important to coarsely select the objects required for answering the question, but the precise regression of their bounding box coordinates is not important.

\section{The impact of information on reasoning requirements}

\begin{figure}[t]
\centering
\includegraphics[width=0.5\textwidth]{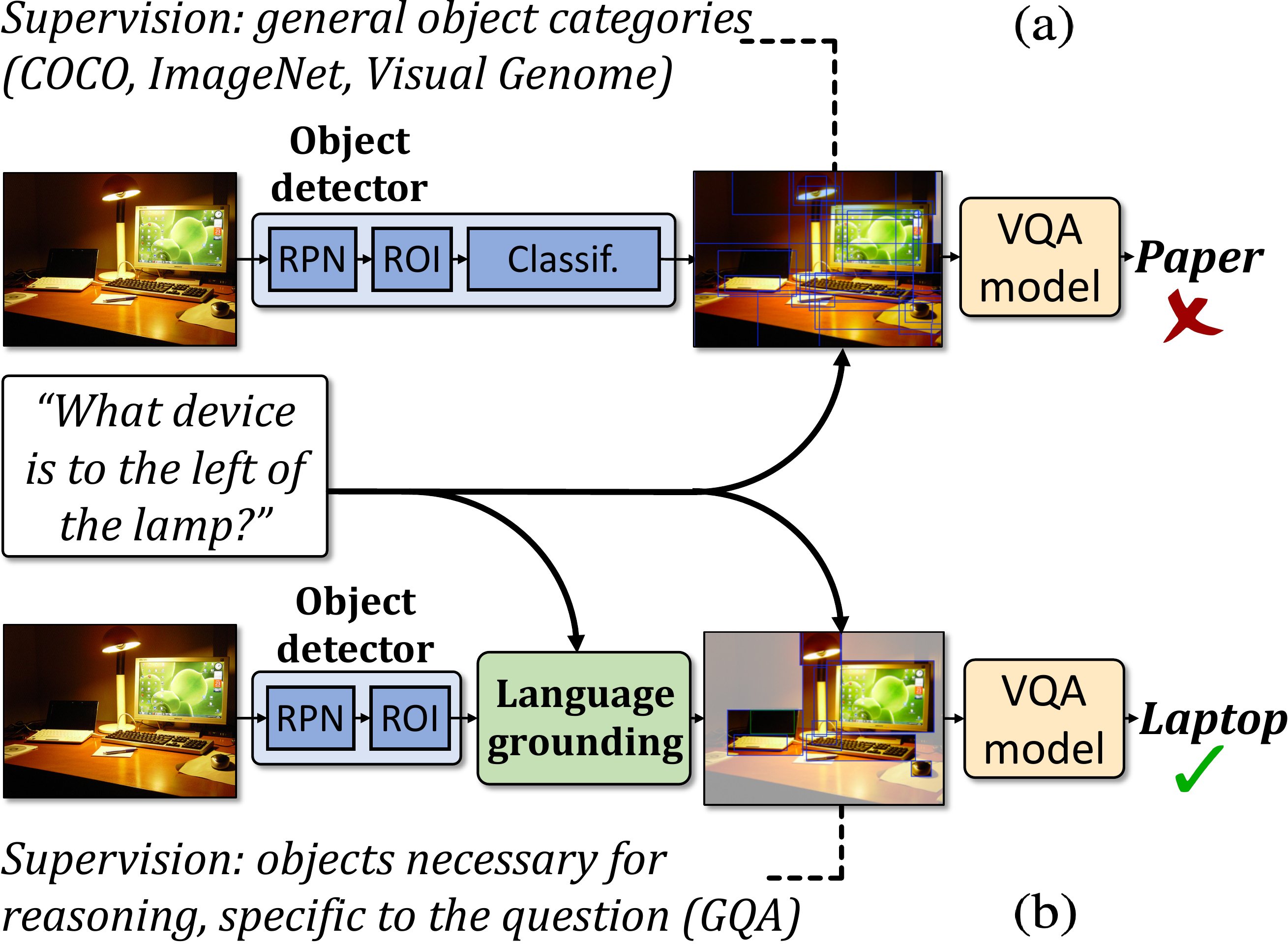}
\caption{\label{fig:teaser}In the context of VQA, we add a task-related top-down process (language grounding) to the initial object detection module. Compared to the classical pipeline (a), we supervise the detector with objects specifically chosen for the input pair (image, question) and show, that the Transformer-based grounding module is capable of retrieving objects mentioned directly or indirectly (b). }
\end{figure}

\noindent
The experiments in section \ref{sec:cvbottleneck} provided evidence for the large impact of high-quality visual input on downstream VQA performance. We now explore the question of the impact of more finegrained information on the visual input. A typical image-question input pair contains a large number of visual objects, but not all objects are required for the reasoning process, i.e. to answer the question. For the example shown in Figure \ref{fig:teaser}, the only objects necessary to answer the question ``\emph{What device is to the left of the lamp}'' are the lamp and the laptop (the answer class). For certain datasets, like the GQA dataset, precise information on the objects required for reasoning is available. We study whether this information can be exploited during training, either through additional supervision, or through a filtering process.

\subsection{Additional supervision of the VQA reasoning model}

\noindent
Pure bottom-up object detectors as currently used in the VQA literature tend to extract the most diverse set of visual objects. As a result, discriminating between useful and unnecessary objects is an important skill, which must be implicitly learnt by VQA models indirectly from classical objectives. We propose to augment the training process with auxiliary supervision, which directly trains this capacity. The underlying hypothesis is that downstream performance will be improved when the model is explicitely guided into recognizing whether a given object is necessary for reasoning.

To this end, we add a classification head to the UpDn VQA model, classifying each object as necessary or not, using the visual inputs from Faster R-CNN. This additional problem can thus be framed as binary classification with a binary cross-entropy loss. The auxiliary loss is summed to the original VQA cross-entropy loss supervised with GT answers to the questions in the dataset.

\myparagraph{Experimental setup} The UpDn VQA model is trained classically, with only a supplementary term to the loss function. The latter is framed as a binary cross entropy loss $\mathcal{L}=\sum^{N}_{i=1} -(y_{i}\log{p_i} + (1-y_i)\log{(1-p_i)})$. Visual objects proposed by the baseline detector (Faster R-CNN pre-trained on Visual Genome used by UpDn) are labelled as positive or negative, based on their IoU with GQA GT BBs.

\myparagraph{Results} Table~\ref{tab:classif_loss} shows results on the GQA balanced validation set after training UpDn on GQA balanced train set. The additional supervision brings a significant boost in performance, showcasing the effect of explicitly incorporating the inductive bias of discriminating necessary objects from unnecessary ones.
\begin{table*}[t]
\centering
{\small
\begin{tabular}{ l c c c c c}
\textbf{Visual inputs from} & \textbf{Classification of necessary objects} &  \textbf{Accuracy} & \textbf{Binary} & \textbf{Open} \\
\hline
Faster R-CNN & $-$ & ${60.24}$ & ${73.57}$ & ${47.75}$\\
Faster R-CNN & \checkmark & ${\textbf{62.21}}$ & ${\textbf{76.86}}$ & ${\textbf{48.48}}$ \\
\end{tabular}
}
\caption{\label{tab:classif_loss}The impact of auxiliary supervision about visual objects necessary for reasoning on GQA  balanced validation set.
}
\end{table*}

\subsection{Language-grounded object selection}
\label{subsec:language_grounding}
\noindent
As an alternative strategy, instead of supervising the VQA reasoning model to discard unnecessary information, we propose to act at the level of the object detector to first only output objects necessary for reasoning, but more importantly, to select objects that would have been missed by a baseline detector without context information about the question.

This is motivated by research in biological vision, which indicates that vision is guided by, both, bottom-up and top-down processes, saliency and the task at hand \cite{BUTDNeuroNature2002,BUTDNeuroScience2007}.
The role of attention in human cognition is central and not only limited to vision, as attention acts as a way to limit bandwidth between different areas of the brain. There is evidence, that even internal attention can be separated into task related processes and processes unrelated to goals \cite{InternalAttention2016}. While current work on VQA acknowledges the important role of bottom-up and top-down processes, their respective roles are clearly divided, and object detection is relegated to a pure bottom-up role \cite{Anderson_2018_CVPR}.
Current VQA approaches rely on bottom-up vision in the form of off-the-shelf object detection models, especially the widely-adopted Faster R-CNN detector \cite{faster_rcnn2015}. Its role is limited to providing bounding boxes and feature embeddings for general object categories, independent of the task at hand and of the question to be answered. They proceed by first predicting a large amount of region proposals, i.e., candidate object instances, which are then individually confirmed or rejected based on the confidence of an object classifier that they belong to any category of its visual vocabulary (Figure~\ref{fig:teaser}a). In the context of VQA (and most other applications requiring object detection), the confirmation and rejection step is necessary. 

Our language-grounded object detector acts on the level of the object classifier of modern two-stage object detectors.
We propose to replace the classifier $\phi_c(\cdot)$ in its role of the region proposal selector by a dedicated language-grounded object selection module $\phi_s$.
Unlike $\phi_c(\cdot)$, the proposed selection module $\phi_s$ takes as input not only the region proposals $\bm{r}$, but also the question $\bm{q}$ asked about the image $\bm{I}$. The module $\phi_s$ serves as ``language grounding'' (i.e., conditioning) for the resulting set of objects $\bm{o}_{LG}$:
\begin{equation}
    \bm{o}_{LG}=\phi_s(\bm{r}(\bm{I}),\bm{q})
\end{equation}
The objective of language grounding is to add a task-related process to object selection after the region proposal stage, avoiding the rejection of relevant objects. A deep neural network combines the encoded question $\bm{q}$ with the region proposals $\bm{r}$ and selects objects, which are considered necessary for reasoning on this input sample. We train the network in a fully supervised way and benefit from the availability of appropriate annotations in the GQA dataset \cite{Hudson2019GQAAN}. For each image-question pair $(\bm{I},\bm{q})$, the dataset contains the set $\bm{o}^*$ of ground truth BBs corresponding to the objects, which are required to answer the question. This GT, provided with GQA dataset, has been calculated from the image scene-graphs, which have been used to procedurally generate the input questions \cite{Hudson2019GQAAN}.

The language grounding module $\phi_s$ is a Transformer-based model, which in similar forms has proven to be effective for vision-language fusion in recent VQA work \cite{Yu_2019_CVPR,Tan2019LXMERTLC,kervadec2019weak}. It takes the full set of proposals $\bm{r}$ extracted by the RPN and the question $\bm{q}$, and produces an output sequence $\bm{o}_{LG} {=} \phi_s(\bm{r},\bm{q})$.
Each input region proposal $\bm{l}\in\bm{r}$ is defined as the concatenation of the $2048$-dimensional feature embedding $\bm{f}_i$ and the corresponding $4$-dimensional BB coordinates $\bm{b}_i$.
The input question $\bm{q}$ is first tokenized using the WordPiece tokenizer \cite{wu2016google}, and then encoded with a standard pre-trained BERT model \cite{devlin-etal-2019-bert} producing a set of $768$-dimensional question embeddings $\bm{q}=\{\bm{q}_0,\bm{q}_1,\mydots,\bm{q}_{N_w-1}\}$ of size $N_w$.
The core architecture is composed of a sequence of intra-modality $t_{-}(\cdot)$ and cross-modality $t_{\times}(\cdot)$ Transformers. See subsection~\ref{subsec:transformers} for a brief presentation of intra and cross-modality Transformers.

The input region proposals $\bm{r}$ are first transformed by $N_r$ intra-modality Transformers to identify relationships between them:
\begin{equation}
    \bm{r}'=t_{-}(\bm{r})\;\;(\times N_r),
\end{equation}
where we omitted parameters (not shared) from the notation. 
Then, the resulting region proposals $\bm{r}'$ and the question embeddings $\bm{q}$ are processed by $N_x$ cross-modality Transformers $t_{\times}$ in order to fuse the information from the two modalities:
\begin{equation}
    \{\bm{r}'',\bm{q}''\}=t_{\times}(\bm{r}',\bm{q})\;\;(\times N_x).
\end{equation}
The final selection scores for each region proposal are calculated by a linear layer, $\phi_s(\bm{l}) {=} \rho ( \bm{W}^s \bm{r}''_{i})$, 
where $\rho$ is the Sigmoid activation.

\begin{figure*}[t]
\centering
\includegraphics[width=1\textwidth]{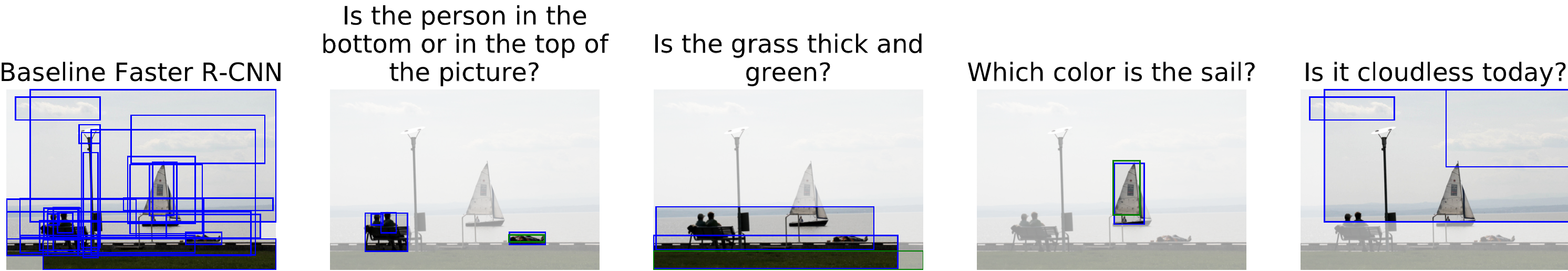}
\includegraphics[width=1\textwidth]{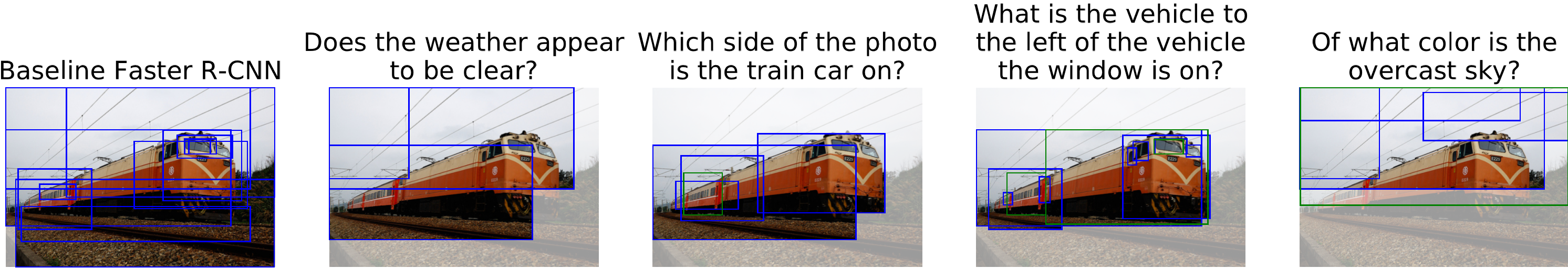}
\includegraphics[width=1\textwidth]{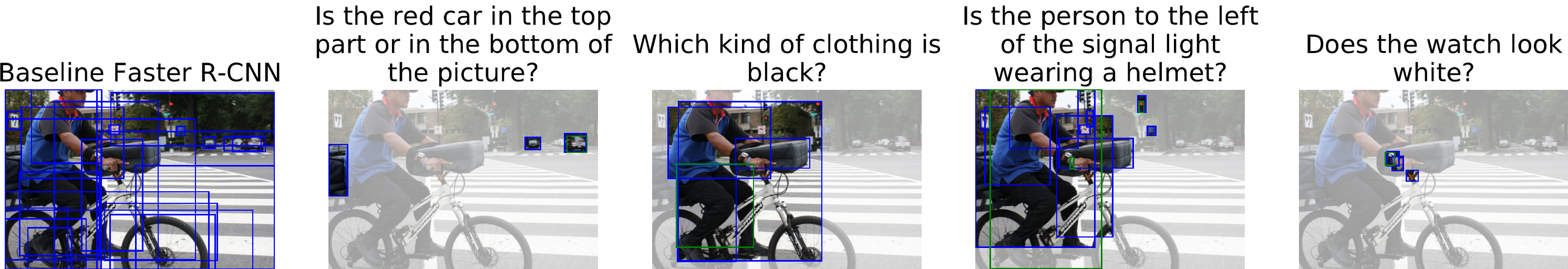}
\caption{Comparison between the visual objects obtained with our language-grounded object selection (columns $2{-}5$) and the purely-visual baseline (column $1$, following \cite{Anderson_2018_CVPR}) that will then be fed to any VQA model. Unlike ours, the baseline does not depend on the question. Blue and green BBs correspond, respectively, to the selected objects and to the GQA GT BBs of the answer objects.}
\label{fig:vizRois}
\end{figure*}

\myparagraph{Experimental setup}
We train the language-grounded object selection module $\phi_s$ separately from the RPN $\phi_r(\cdot)$, to classify each input region proposal as ``positive'' or ``negative'' given a grounding question. We create the Ground Truth labels for predicted BBs by matching them with the GT GQA boxes and keeping correspondences with Intersection over Union (IoU) above a  threshold $\theta_{IoU}$. Training minimizes the binary cross-entropy loss
$\mathcal{L}=\sum^{N}_{i=1} -(y_{i}w_{+}\log{p_i} + (1-y_i)w_{-}\log{(1-p_i)})$.
To cope with an imbalanced dataset ($1$\% of positives), we weight false positives and false negatives with weights $w_{+}$ and $w_{-}$, respectively.

For a fair comparison with the mainstream object selection approach in VQA \cite{Anderson_2018_CVPR}, \textit{cf.} Section~\ref{subsec:updn}, which serves as a primary baseline for our work, we do not refine the selected RPN's region proposals apart from concatenating the BBs coordinates to the respective feature vectors (\textit{i.e.,} following the notation of Section~\ref{subsec:object_detection}, if a region $\bm{l}\in\bm{r}$ is selected, then $\bm{l}\in\bm{o}$).
Before performing object selection, we firstly apply a Non-Maximum Suppression (NMS) with the IoU threshold of $\theta_{NMS_1}$ on the outputs of RPN using its "objectness" as confidence score. A more strict NMS with the IoU threshold of $\theta_{NMS_2}$ is then performed on the selected objects using the respective selection scores.
The NMS thresholds and other training and design hyperparameters have been chosen via validation and set as the following: $\theta_{NMS_1}=0.7$, $\theta_{NMS_2}=0.4$, $N_r=3$, $N_x=3$, $\theta_{c}=0.2$, $\theta_{s}=0.5$, $\theta_{IoU}=0.5$, $w_{+}=40$ and $w_{-}=1$.

The language-grounded selection model is trained on the balanced train split of the GQA dataset and validated on the balanced validation split. It is trained during 5 epochs with a batch size of 32 using the SGD optimizer. The learning rate is set to $2e^{-3}$ and multiplied by $0.1$ every 3 epochs. 

\myparagraph{Language-grounded vs. purely-visual detection}
\noindent
We first evaluate the language-grounded object selection module on its pure detection performance by comparing it with the standard RPN region proposal ranking used in~\cite{Anderson_2018_CVPR}. We present qualitative results of  language-grounded object detection vs. classical bottom-up detection in Figure~\ref{fig:vizRois}:
the language grounded approach focuses on question-relevant objects by nicely matching the GT BBs, and
also filters out unnecessary objects.
More importantly, the $2$ right-most images of the 3rd row also highlight the capacity of the approach to select useful objects (the signal lights and the man's watch) that are ignored by the baseline.
Figure~\ref{fig:vizRois} also provides evidence that language-grounding is able to retrieve objects, which are only indirectly mentioned.
For instance, it detects the sky, of the image when asked ``Does the weather appear to be clear?''

It is obvious, that the connection between VQA performance and object selection quality is not direct. One may argue that a VQA question can be potentially answered even if not all necessary objects are detected.
For example, a missed object can be covered by parts of several detected objects (that is why the purely-visual baseline usually requires many objects to be selected for VQA).
As a result, we next study the impact of language-grounded detection on the downstream VQA performance.

\begin{table*}[t]
\centering
{\small
\begin{tabular}{c c ccc ccc}
\textbf{Visual inputs from}  & \textbf{\# objects} &
\multicolumn{3}{c}{\textbf{--- Test-dev ---}} &
\multicolumn{3}{c}{\textbf{--- Test ---}} 
\\
&&
Accuracy & Binary & Open  &
Accuracy & Binary & Open   \\
\hline
Faster R-CNN & $36$ & $53.30$ & $70.23$ & $38.93$  & $54.04$ & $69.75$ & $40.17$\\ 
Faster R-CNN + LG & $12.23$ & $\textbf{54.27}$ & $\textbf{71.78}$ & $\textbf{39.42}$ & $54.25$ & $70.59$ & $39.82$ \\
\end{tabular}
}
\caption{\label{tab:results_gqa}Downstream performance of the UpDn model when trained on GQA train+val balanced set and evaluated using visual inputs either selected by the language-grounded module or the standard purely-visual selection pipeline. LG denotes language grounding}
\end{table*}

\begin{table}[t]
\centering
{\small
\begin{tabular}{ p{30mm}  c c }
\multicolumn{3}{c}{} \\
\toprule
\textbf{Visual inputs from} & \textbf{\# objects} & \textbf{Accuracy}  \\
\midrule
Faster R-CNN & $10$ & $50.86$ \\
Faster R-CNN & $20$ & $52.85$\\ 
\hline
Faster R-CNN + LG & $12.23$ & $\textbf{54.27}$ \\
\bottomrule
\end{tabular}
}
\caption{\label{tab:nb_objects} Comparing downstream performance of the UpDn model with and without the language-grounded module while limiting the number of objects. We observe that, unlike the baseline, the language-grounded module allows reaching a high accuracy with only few objects. Results on GQA test-dev.}
\end{table}

\begin{figure*}[t] \centering
    \includegraphics[width=\textwidth]{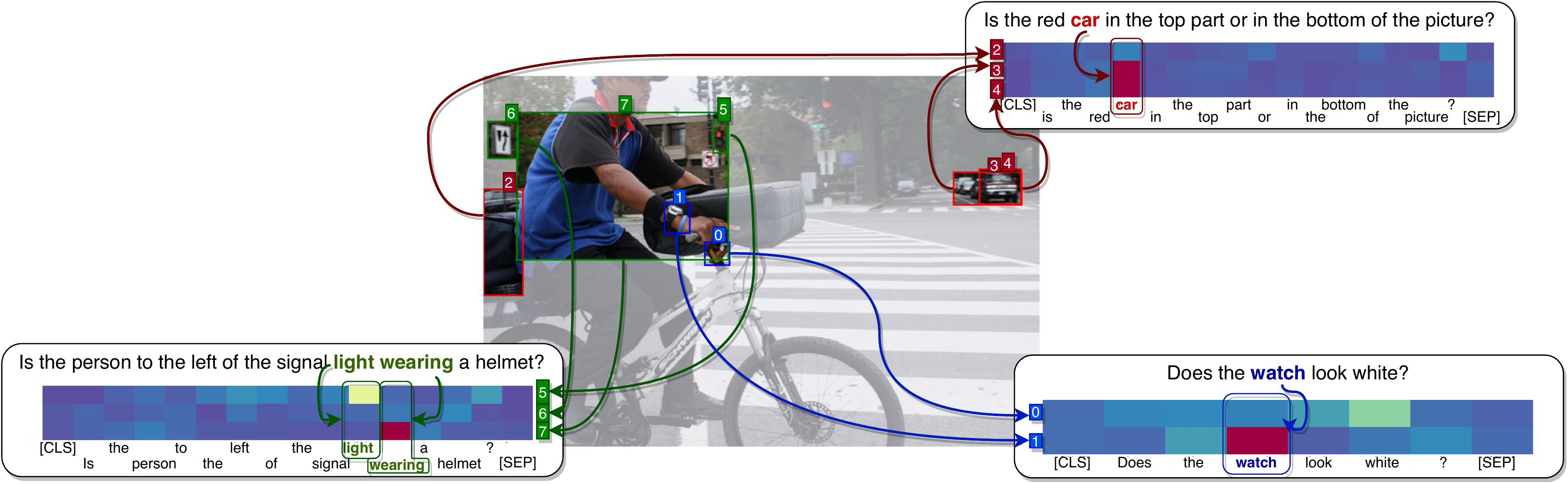}
    \caption{\label{fig:attentionmaps} Visualization of cross-modality attention. Attention coefficients between the image objects and the question tokens are displayed as heatmaps for several questions. Textual tokens matching the semantic content of the bounding boxes are more strongly attended. Regions within the image which are not related to a given question generate a uniform distribution of attention coefficients over its textual tokens}
\end{figure*}

\myparagraph{Impact of language-grounded selection on VQA}
\noindent
The UpDn VQA model is trained on the GQA balanced train+val set and evaluated on GQA test-dev (used for validation) and test sets, using visual inputs either selected by the language-grounded module (\textit{Faster R-CNN + LG}) or using the standard purely-visual selection pipeline as in~\cite{Anderson_2018_CVPR} (\textit{Faster R-CNN}). Obtained results are shown in Table~\ref{tab:results_gqa}. As we can see, the performance improves quite significantly on the test-dev set when feeding visual inputs selected by the language-grounding model. On the test set, however, the gain does not seem to be significant.

We also report in Table~\ref{tab:results_gqa} and \ref{tab:nb_objects} the mean number of selected objects per question with the language-grounded module, which is about $3$ times less than the standard number of objects used in VQA with the Faster R-CNN detector from~\cite{Anderson_2018_CVPR}. This confirms that the studied VQA approaches based on this detector are fed with a not negligible amount of unnecessary input objects, highlighting the advantage of a task-related ranking of visual objects against purely bottom-up ranking of BBs as done in baseline detectors.

\myparagraph{Origins of VQA gains}
We study the question whether the VQA performance gain of language-grounding on the test-dev set is due to filtering of question-irrelevant objects, or, on the contrary, to the inclusion of additional necessary objects w.r.t. the baseline.
Below, we demonstrate that both reasons partly contribute to the performance improvement.
In Table~\ref{tab:ablation}, we compare the performance of the UpDn model on the set of objects detected by the baseline Faster R-CNN without ($1^{st}$ row) and with ($2^{nd}$ row) adding supplementary objects, which had been missed by Faster R-CNN, but detected by our language-grounded detector. Additional objects (\textit{Ours} in $2^{nd}$ row) were selected based on the threshold $\theta_{s}=0.5$ on the score from our selection module but, unlike in the introduced method ($3^{rd}$ row), no NMS was then applied. Instead, only objects with an IoU under $\theta_{IoU}=0.5$ with all objects from Faster-RCNN were kept. 

The difference of about $+1$ point between rows $1$ and $2$ demonstrates that adding supplementary useful objects helps the baseline and testifies that our approach detects essential objects for reasoning  missed by the baseline.
On the other hand, the difference of about $+0.25$ point between the $2^{nd}$ and the $3^{rd}$ row of Table~\ref{tab:ablation} provides evidence that removing unnecessary question-irrelevant and redundant objects simplifies VQA reasoning and improves performance.

\begin{table}[t]
\centering
{\small
\begin{tabular}{ p{30mm}  c c c c }
\multicolumn{4}{c}{} \\
\toprule
\textbf{Visual inputs from} & \textbf{Accuracy} & \textbf{Binary} & \textbf{Open} \\
\midrule
Faster R-CNN & ${53.10}$ & ${69.49}$ & ${39.19}$ \\
Faster R-CNN $\cup$ LG  & ${\textbf{54.16}}$ & ${\textbf{71.40}}$ & ${\textbf{39.53}}$ \\
\hline
LG & ${\textbf{54.39}}$ & ${\textbf{71.92}}$ & ${39.34}$\\
\bottomrule
\end{tabular}
}
\caption{\label{tab:ablation}Difference in performance of the UpDn VQA model on the set of objects detected by the baseline Faster R-CNN without ($1^{st}$ row) and with ($2^{nd}$ row) adding supplementary objects which are missed by Faster R-CNN, but detected by our approach (noted by Faster R-CNN $\cup$ LG). As a reminder, the performance of UpDn when trained and evaluated with only the objects from our language-grounded object detector is provided in the last row. Results on GQA test-dev.}
\end{table}

\myparagraph{Visualization of object-word matching}
Finally, the cross-modality attention map of the language-grounded module is visualized in Figure~\ref{fig:attentionmaps}, illustrating that tokens in the question matching the semantic content of the given BBs are more strongly attended, which makes the method interpretable.
We also find encouraging that regions unrelated to asked question (e.g., BB number 6 in Figure~\ref{fig:attentionmaps}) generate a uniform distribution of attention coefficients over all textual tokens.

\section{Discussion and conclusions}
\label{sec:conclusion}

We presented an in-depth study of the vision-bottleneck in VQA, namely the impact of both quantity and quality of visual objects extracted from input images, and used to reason when answering questions. We then experimented with two ways to incorporate information about objects necessary for reasoning, as an auxiliary supervision of the VQA reasoning system, and directly in the object detector to study the effect of a task-related ranking of object BBs.

Our findings validate recent work \cite{ReasoningCVPR2021}, \cite{ReasoningNeurIPS2021}, highlighting the central role of vision in VQA, and thus the importance of improving the extraction of visual tokens in this task. More specifically, we showed that selecting the right visual objects necessary for answering the question is critical, confirming the importance of the quality of selected BBs, not in terms of their perfect coordinates but rather the latent code associated with the objects of interest. We also demonstrated that a VQA reasoning system can be overwhelmed when fed with too many objects as input, highlighting the importance of the recall in object selection.

When augmenting the training of a reasoning VQA model with an auxiliary supervision aiming at quantifying its capacity to distinguish between objects necessary and useless regarding the asked question, downstream performance improves significantly. Such inductive bias about the importance of weighting the contribution of given inputs is thus an interesting way to guide VQA models, while being relatively simple to implement, unlike more involved methods. When provided with the right supervision, as proposed in the GQA dataset, this additional loss term could be a nice addition to other contributions when trying to maximize VQA accuracy. Future interesting work could also target a similar direction, with less or even no explicit supervision about objects necessary for reasoning.

Finally, when acting directly at the level of the detector, a Transformer-based model is able to filter out irrelevant objects regarding the question, and to select challenging objects within the image. An interesting point to note is its surprising effectiveness to select visual objects that are indirectly mentioned in the question. When ranking objects fed to VQA systems based on the studied task-related  score, overall performance is at least as good as with objects from a baseline detector, but while relying on a third of visual inputs in average. This showcases the lack of relevance of classical object detection ranking when considering VQA. Grounding object detectors with language can thus be considered as a promising direction, that could lead to an even more complete fusion of vision and language modalities in future work.

\textbf{Acknowledgements ---} C. Wolf acknowledges support from ANR through grant ``\emph{Remember}'' (ANR-20-CHIA-0018).

{\small
\bibliographystyle{ieee_fullname}
\bibliography{refs}
}

\end{document}